\documentclass[letterpaper, 10 pt, conference]{ieeeconf}  
\usepackage{censor}
\StopCensoring 

\IEEEoverridecommandlockouts                              

\makeatletter
\let\NAT@parse\undefined
\makeatother

\usepackage{graphics} 
\usepackage{amsmath} 
\usepackage{amssymb}  
\usepackage[hyphens]{url}
\usepackage{hyperref}
\usepackage[hyphenbreaks]{breakurl}
\usepackage{multirow}
\usepackage{xcolor}
\usepackage{tikz}
\usetikzlibrary{calc}

\definecolor{colX}{RGB}{51,34,136}   
\definecolor{colZ}{RGB}{204,102,0}   
\definecolor{colA}{RGB}{17,119,51}   
\definecolor{colB}{RGB}{153,52,4}    
\definecolor{colC}{RGB}{0,102,153}   
\definecolor{colS}{RGB}{0,0,0}       
\definecolor{colD}{RGB}{230,159,0}   
\definecolor{colE}{RGB}{86,180,233}  
\definecolor{colF}{RGB}{213,94,0}    
\definecolor{colG}{RGB}{0,158,115}   
\definecolor{colH}{RGB}{240,228,66}  
\definecolor{colI}{RGB}{204,121,167} 

\newcommand\submittedtext{%
	\scriptsize This work has been submitted to the IEEE for possible publication. Copyright may be transferred without notice, after which this version may no longer be accessible.}

\newcommand\submittednotice{%

		\begin{tikzpicture}[remember picture,overlay]
			\node[anchor=south,yshift=0pt] at (current page.south) 	{\fbox{\parbox{\dimexpr0.95\textwidth-\fboxsep-\fboxrule\relax}{\submittedtext}}};
		\end{tikzpicture}%
}

\graphicspath{ {images/} } 

\title{\LARGE \bf From Transportation to Manipulation: \\ Transforming Magnetic Levitation to Magnetic Robotics}

\author{\censor{Lara Bergmann$^{1}$, Noah Greis$^{1}$,  and Klaus Neumann$^{1,2}$}
\xblackout{\thanks{*This work was not supported by any organization}
\thanks{$^{1}$Lara Bergmann, Noah Greis, and Klaus Neumann are with the CITEC, Faculty of Technology, Bielefeld University, 33619 Bielefeld, Germany
        {\tt\small {\{lara.bergmann, noah.greis, klaus.neumann\}} @uni-bielefeld.de}}%
\thanks{$^{2}$Klaus Neumann is with Fraunhofer IOSB-INA, 32657 Lemgo, Germany}}%
}

\begin{document}
\maketitle
\thispagestyle{empty}
\pagestyle{empty}
\submittednotice
\begin{abstract}
Magnetic Levitation (MagLev) systems fundamentally increase the flexibility of in-machine material flow in industrial automation. Therefore, these systems enable dynamic throughput optimization, which is especially beneficial for high-mix low-volume manufacturing. Until now, MagLev installations have been used primarily for in-machine transport, while their potential for manipulation is largely unexplored. This paper introduces the \mbox{\emph{6D-Platform MagBot}}, a low-cost six degrees of freedom parallel kinematic that couples two movers into a composite robotic platform. Experiments show that the \mbox{\emph{6D-Platform MagBot}} achieves sub-millimeter positioning accuracy and supports fully autonomous pick up and drop off via a docking station, allowing rapid and repeatable reconfiguration of the machine. Relative to a single mover, the proposed platform substantially expands the reachable workspace, payload, and functional dexterity. By unifying transportation and manipulation, this work advances Magnetic Levitation towards \emph{Magnetic Robotics}, enabling manufacturing solutions that are more agile, efficient, and adaptable.
\end{abstract}
\section{INTRODUCTION}
\begin{figure}
	\centering
    \includegraphics[width=\linewidth]{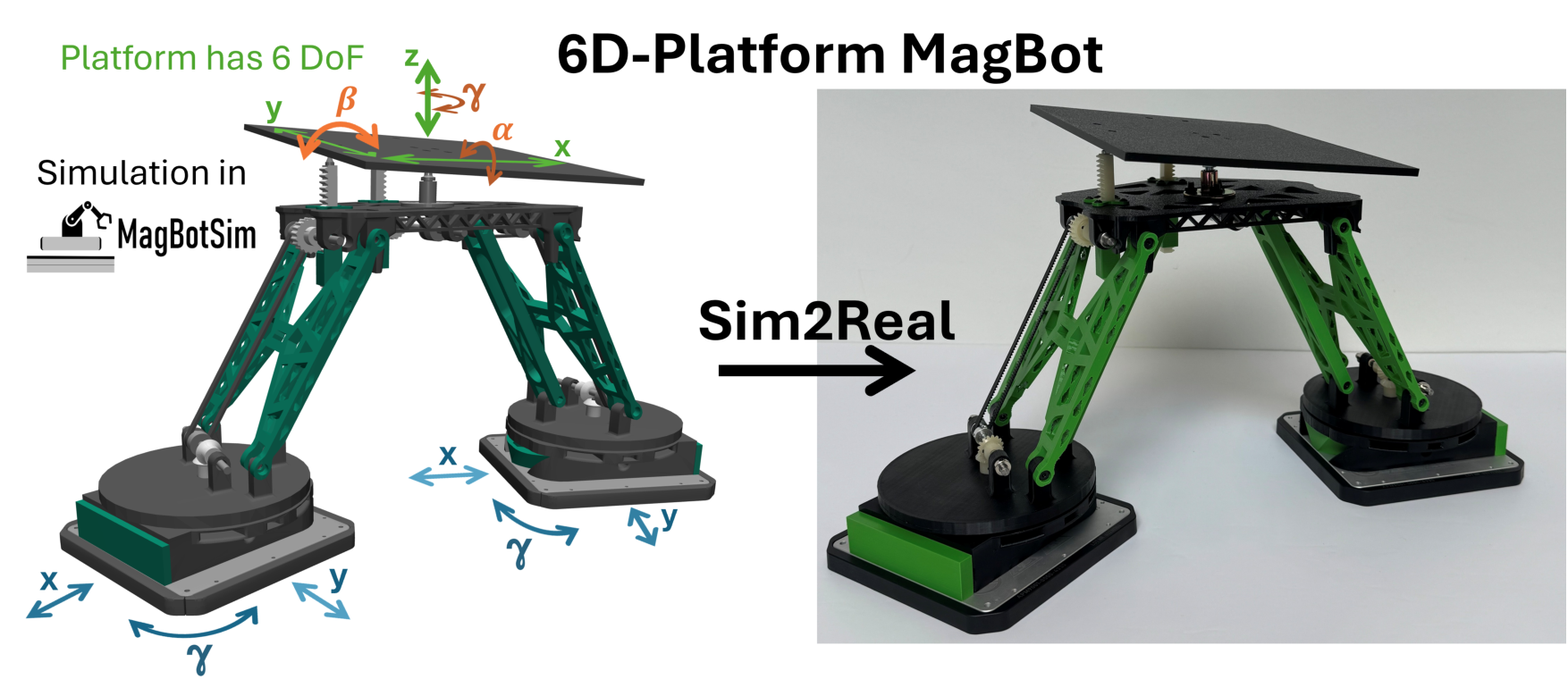}
    \caption{\emph{6D-Platform MagBot} in simulation (left) and reality (right). The platform has 6 DoF and is controlled by the x-,y-, and $\gamma$-axes of the movers.}
    \label{fig_sim2real}
    \vspace{-0.3cm}
\end{figure}

\noindent
By transforming in-machine material flow, Magnetic Levitation (MagLev) provides the inherent flexibility required for high-mix low-volume industrial automation. These MagLev systems are composed of static motor modules, called \emph{tiles}, and \emph{movers}, consisting of multiple Halbach arrays \cite{lu_6d_2012}. During operation, the tiles emit electromagnetic fields, thereby controlling the movers very precisely in six dimensions (see Fig.~\ref{fig_visual_abstract}). The modular tiles can be arranged in nearly any planar configuration, and movers of different sizes can be flexibly added or removed from the tile surface, making such systems highly reconfigurable. Several MagLev systems for industrial applications exist, such as XBot (Planar Motor), XPlanar (Beckhoff Automation), ACOPOS 6D (B\&R Industrial Automation), and ctrlX $\text{FLOW}^{\text{6D}}$ (Bosch Rexroth). These agile transport systems have inherent, though currently underutilized, capacities for manipulation. Although the workspace of the movers varies slightly depending on the MagLev system used, the workspace is generally very limited, restricting the manipulation capabilities of the system. Therefore, some attempts have been made to compensate for these limitations, however, with a limited dimensionality of the workspace~\cite{Lu2020,Lu2022}. To expand the workspace in all dimensions, payload and functional dexterity compared to a single mover, we present the \emph{6D-Platform MagBot} (see Fig.~\ref{fig_sim2real}), a low-cost parallel kinematic with six degrees of freedom (DoF) that couples two movers into a composite robotic platform. The \emph{6D-Platform MagBot} is designed to be compatible with all aforementioned MagLev systems. Additionally, we developed a docking station where movers can autonomously drop off or pick up the \mbox{\emph{6D-Platform MagBot}}, shown in Fig.~\ref{fig_visual_abstract}. Therefore, a machine can be easily reconfigured, which shortens changeovers: adapting to new products or variants becomes faster and requires less engineering time.
By unifying transportation and manipulation in one single paradigm for the domain of in-machine material flow, we transform Magnetic Levitation towards Magnetic Robotics. Hence, Magnetic Robotics enables manufacturing systems that are simultaneously more agile, efficient, and flexible. In summary, our work makes the following contributions:
\begin{figure}
	\centering
    \includegraphics[width=\linewidth]{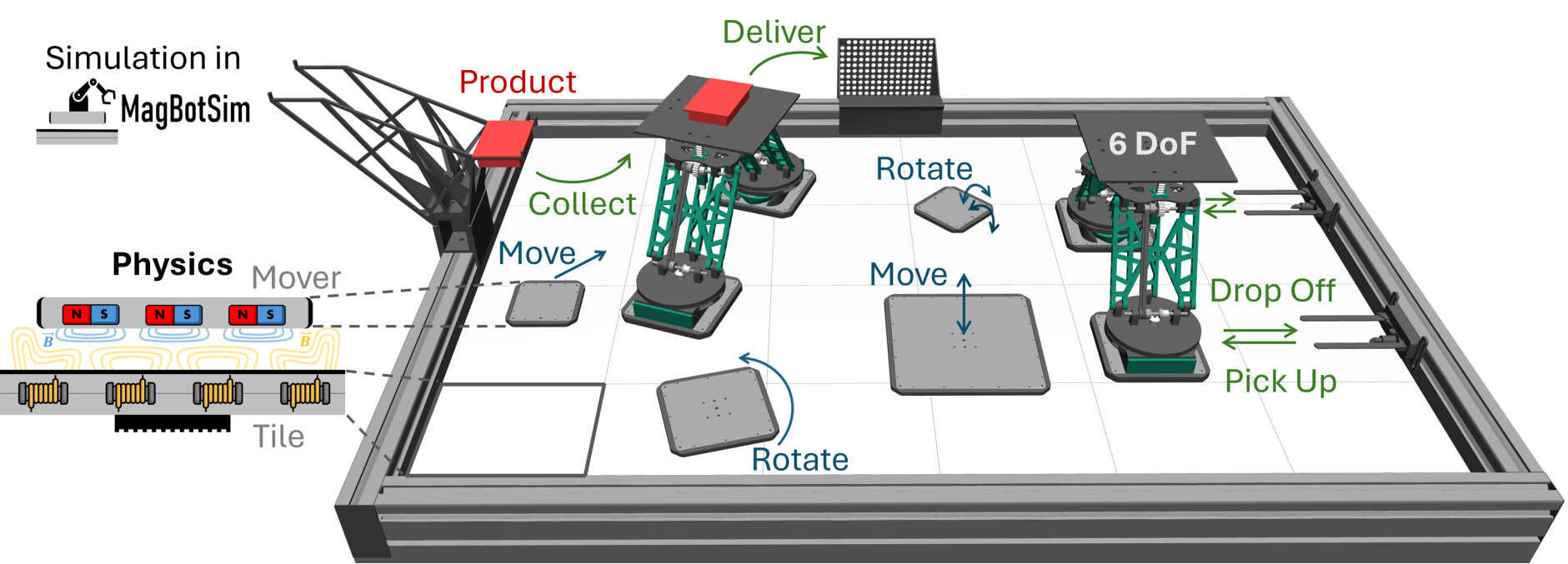}
        \caption{Schematic illustration of Magnetic Robotics. The \emph{6D-Platform MagBot} couples two movers and has a docking station for reconfigurability.}
    \label{fig_visual_abstract}
\end{figure}
\begin{itemize}
    \item Designed and built the \emph{6D-Platform MagBot}, including a dedicated docking station for reconfigurability.
    \item Developed and implemented an inverse kinematics controller for 6D motion control.
    \item Created a physics-based simulation of the \emph{6D-Platform MagBot} integrated into MagBotSim~\cite{bergmann_magbotsim_2025}.
    \item Conducted rigorous experiments to quantify the platform's precision and evaluate its applicability.
\end{itemize}
Videos, CAD-files, a component list, and assembly instructions for the \emph{6D-Platform MagBot} are available at: \url{https://sites.google.com/view/6d-platform-magbot?usp=sharing}
\section{RELATED WORK}
\noindent
\textbf{Magnetic Levitation.} 
Latest works on MagLev focus on hardware development and low-level control of a single mover~\cite{lu_6d_2012,trakarnchaiyo_design_2023,hartmann_end--end_2025}. From a \emph{Magnetic Robotics} perspective, trajectory planning that considers system-specific requirements, such as high dynamics and smoothness, is rarely studied yet~\cite{pierer_von_esch_sensitivity-based_2025,janning_conflict-based_2025}. Compared to medical applications, where trajectory planning and object manipulation with magnetic microrobots are investigated~\cite{isitman_trajectory_2025,sallam_autonomous_2024,song_motion_2023}, we consider MagLev systems that are larger, stronger, and used in the industrial domain. Therefore, approaches for medical use cases cannot be directly transferred. To enable the development of approaches for industrial MagLev systems, we introduced MagBotSim~\cite{bergmann_magbotsim_2025}, a physics-based simulation for \emph{Magnetic Robotics}, and showed that MagLev systems can manipulate objects by pushing. In addition, we suggested collaborative object manipulation, explicitly mentioning the coupling of movers by a parallel kinematic, as a future research direction. This work builds on these suggestions by introducing the \emph{6D-Platform MagBot} as the first reconfigurable 6-DoF kinematic mechanism that couples two movers.\\\\
\textbf{6-DoF Kinematics.} The Steward Platform~\cite{stewart_platform_1965}, the HEXA robot~\cite{pierrot_towards_1991} (a 6-DoF variant of the DELTA robot~\cite{clavel_conception_1991}), and the Hexapteron~\cite{seward_new_2014} are well-known 6-DoF kinematic mechanisms. However, they are all mounted on a stationary base. The Triple Scissor Extender~\cite{gonzalez_triple_2016} is another 6-DoF kinematic mechanism mounted on linear stages that are limited in length and thus limit the workspace of the robot. In contrast, our \emph{6D-Platform MagBot} is reconfigurable through a docking station and mounted on two movers that are non-stationary. In fact, the workspace of the movers, and thus of the \emph{6D-Platform MagBot}, can be arbitrarily large in x- and y-directions, as it is only limited by the number of tiles in each direction. Other 6-DoF kinematics, such as 3-PPRS~\cite{liu_parallel_2014} or 3-PPSR~\cite{liu_parallel_2014}, consist of three serial chains mounted on three bases that move in x- and y-directions relative to each other. Since the movers of a MagLev system can not only move in x- and y-directions, but also rotate around the z-axis of the mover, we can eliminate the third base and its associated chain, which simplifies the inverse kinematics controller and the kinematic mechanism itself.\\\\
\textbf{Reconfigurable Robots.} Reconfigurable robots, such as PuzzleBot~\cite{yi_puzzlebots_2021}, FreeBOT~\cite{liang_freebot_2020}, 3D M-Blocks~\cite{romanishin_3d_2015}, Slimebot~\cite{shimizu_amoeboid_2009}, Swarm-Bot~\cite{mondada_swarm-bot_2004}, or M-TRAN III~\cite{kurokawa_distributed_2008}, typically consist of several building blocks (modules) that can either act autonomously or change their physical configuration by coupling and decoupling. Since each module can move on its own, it must contain electronics, e.g. processors, actuators, and components for communication with other modules. In contrast, our \emph{6D-Platform MagBot} does not contain any additional electronics, as the actuators are already provided by the MagLev system in the form of the movers. Additionally, no wireless communication between MagBots is required, as the position of all movers is provided by the MagLev system, and therefore precisely and globally available in each control cycle. Furthermore, the reconfigurability of the \emph{6D-Platform MagBot} refers to the adaptation of a machine, rather than a shape change or a coupling of multiple modules. Therefore, our \emph{6D-Platform MagBot} together with the MagLev system is a special kind of reconfigurable robot.
\section{DESIGN DECISIONS}
\begin{figure}
	\centering
    \includegraphics[width=\linewidth]{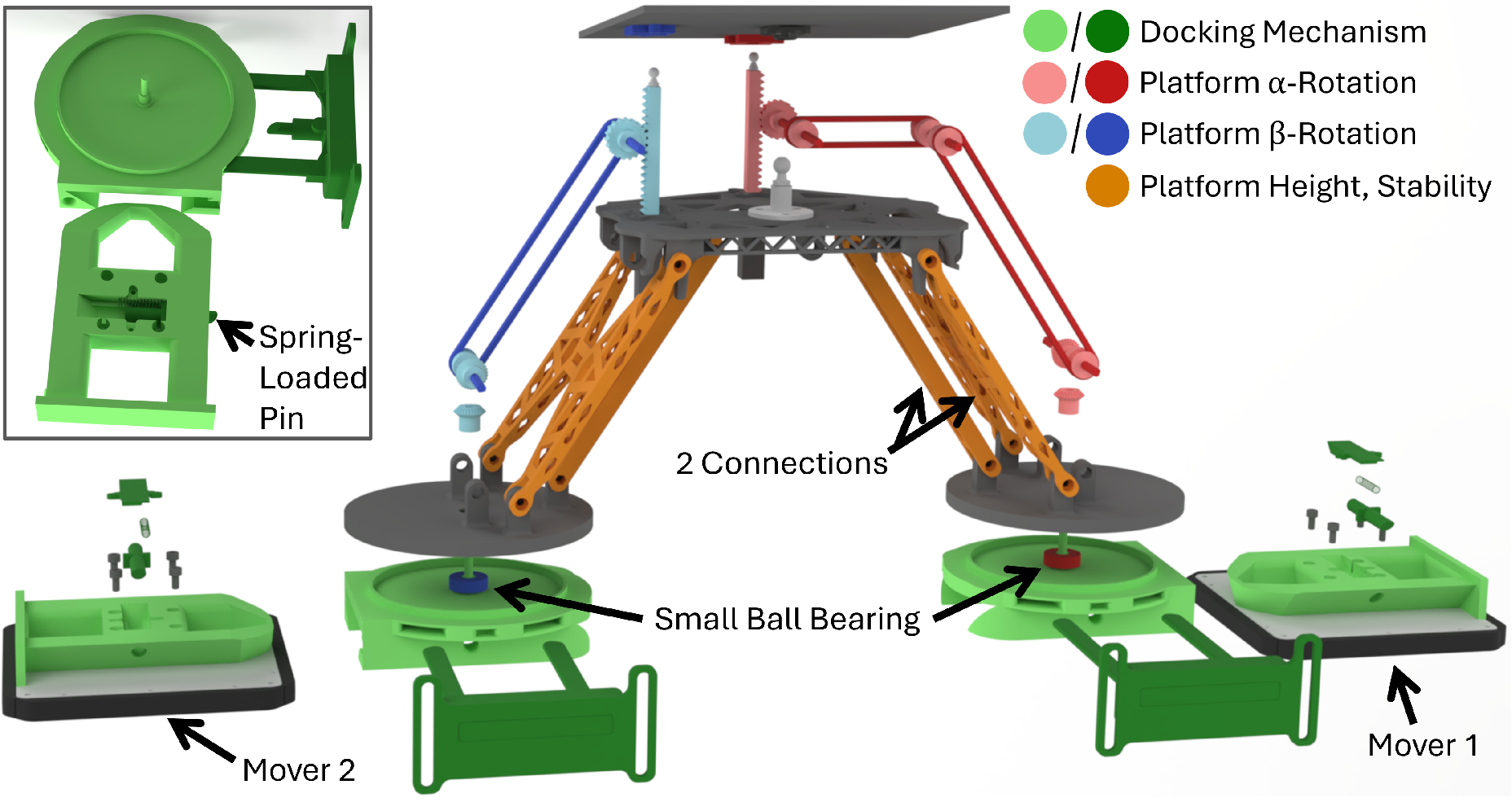}
        \caption{Exploded view of the \emph{6D-Platform MagBot} and visualization of the docking mechanism (top left). Components for docking, the platform's $\alpha$- and $\beta$-rotations, height, and stability are marked in different colors.}
    \label{fig_explosion_view}
\end{figure}
\noindent The \emph{6D-Platform MagBot} is designed with respect to the following conditions:
\begin{itemize}
    \item Small footprint (coupling only two movers)
    \item Self-sufficiency (no batteries, no actuation)
    \item Reconfigurability (pick up/drop off docking station)
    \item Compatibility (with all above-named MagLev systems)
    \item Affordability (low-cost, lightweight, 3D-printed)
    \item Robustness (high dynamics, durability, repairability)
\end{itemize}
Throughout this paper, $\alpha,\beta$, and $\gamma$ denote the rotations around the x-,y-, and z-axes, respectively, as shown in Fig.~\ref{fig_sim2real}.
\subsection{6-DoF Platform}
\noindent
Since our goal is to use the movers of the MagLev system as actuators for the \emph{6D-Platform MagBot} defined as a parallel kinematic, no additional actuation is required. More specifically, the platform is controlled by the x-,y-, and $\gamma$-axes of the movers. Repositioning the movers in x- and y-directions is required to adjust the x-, y-, and z-positions of the platform, where the platform's z-position is changed by altering the distance between the two movers. Mechanically, the movers are coupled by an isosceles trapezoid (see Fig.~\ref{fig_explosion_view}), where the components marked in orange are connected to the gray parts by revolute joints. We use two connections depicted in orange on each side between the mover and platform (see Fig.~\ref{fig_explosion_view}) to stabilize the platform and prevent it from tilting. The $\alpha$- and $\beta$-rotations of the platform are coupled with the $\gamma$-rotations of the movers (mover 1: $\alpha$-rotation, mover 2: $\beta$-rotation). Mechanically, we use ball bearings marked in dark blue and red between the parts depicted in gray and light green at the bottom of Fig.~\ref{fig_explosion_view}. When connecting the light green and gray components, a shaft from the light green part goes through the gray part and is equipped with a bevel gear (light blue/light red). A second bevel gear on each side rotates the mover's $\gamma$-rotation by $90^\circ$, thereby driving the lower shaft. A toothed belt is used to transmit the rotation to the second shaft. Spur gears are mounted on these shafts that are connected with toothed racks. By moving the toothed racks vertically up and down, the top platform gets rotated in $\pm\alpha$ and $\pm\beta$ depending on the mover's rotational direction. All components that are relevant to transmit the mover's $\gamma$-rotation to the platform are marked in red (platform $\alpha$-rotation) and blue (platform $\beta$-rotation) in Fig.~\ref{fig_explosion_view}. The general idea of transmitting the mover's $\gamma$-rotation is similar for $\alpha$ and $\beta$. The main difference is that we need two toothed belts for $\alpha$ and only one for $\beta$ to rotate the platform correctly. The platform is supported using three ball joints: one located at the center of the platform, while the remaining two are mounted on the toothed racks. The latter can move linearly to enable the rotation of the platform. The platform's $\gamma$-rotation can be changed by moving the movers in x- and y-directions on a circle. However, this also requires compensation for the $\alpha$- and $\beta$-rotations of the platform using the $\gamma$-rotation of the movers due to the mechanical coupling.
\subsection{Docking Mechanism}
\noindent
To make a machine easily reconfigurable, we designed a docking station that allows to autonomously drop off or pick up the \emph{6D-Platform MagBot}. The docking mechanism is a self-locking rail system shown in green in Fig.~\ref{fig_explosion_view}. The rail system is a two-part dovetail joint, where the first component, being screwed onto a mover, moves into the second one that is a part of the \emph{6D-Platform MagBot} (see Fig.~\ref{fig_explosion_view} top left). The component on the mover has a small spring-loaded pin inside. The mechanism locks as soon as the pin snaps into the hole of the MagBot's component. To unlock the mechanism and autonomously drop off the \emph{6D-Platform MagBot}, we designed a docking station marked in dark green (see Fig.~\ref{fig_explosion_view}), which slides into the recesses of the MagBot's docking component and has a counter pin to push back the spring-loaded pin, thereby unlocking the aforementioned MagBot and mover components. The forks of the docking station are reinforced with stainless steel plates and can carry the \emph{6D-Platform MagBot}. We would like to emphasize that the docking mechanism is designed such that it can not only be used for the \emph{6D-Platform MagBot}, but also for other MagBots or tools that are to be mounted on a mover, ensuring the reconfigurability of a machine.
\subsection{Compatibility with Different MagLev Systems}
\noindent
The \emph{6D-Platform MagBot}, including the docking station, is compatible with all aforementioned MagLev systems without any major modifications. Only the screw holes in the component that is screwed onto the mover may need to be adjusted, as the holes in the movers might be placed differently when using another MagLev system.
\subsection{Cost and Weight}
\noindent
Since the movers have a limited payload, the main components of the \emph{6D-Platform MagBot} are 3D-printed with $20\%$ infill using PETG filament to ensure stability. In addition, all other components are made of plastic, except for screws, ball bearings, and ball heads. We decided to use small ball bearings in the base parts of the \emph{6D-Platform MagBot} (see Fig.~\ref{fig_explosion_view}) to reduce weight. Thus, one \emph{6D-Platform MagBot} weighs approximately $1.09\,$kg. Note that this weight is distributed across two movers. Furthermore, the \emph{6D-Platform MagBot} is low-cost, as one MagBot costs approximately $240\,$USD. Using 3D-printed and plastic components, as well as the small ball bearings, comes at a price in terms of precision. However, we show that the positioning accuracy of the platform is in the sub-millimeter/sub-degree range except for simultaneous $\alpha$- and $\beta$-rotations (see Sec.~\ref{sec_sa_accuracy}, \ref{sec_ma_accuracy}), which is sufficient for most industrial applications. Nevertheless, future work should investigate building more precise versions of the \emph{6D-Platform MagBot} using different materials, although this would likely make the MagBot more expensive.
\section{INVERSE KINEMATICS CONTROLLER}
\begin{figure}
	\centering
	\begin{tikzpicture}
		\node[anchor=south west, inner sep=0] (img) at (0,0)
		{\includegraphics[width=0.8\linewidth]{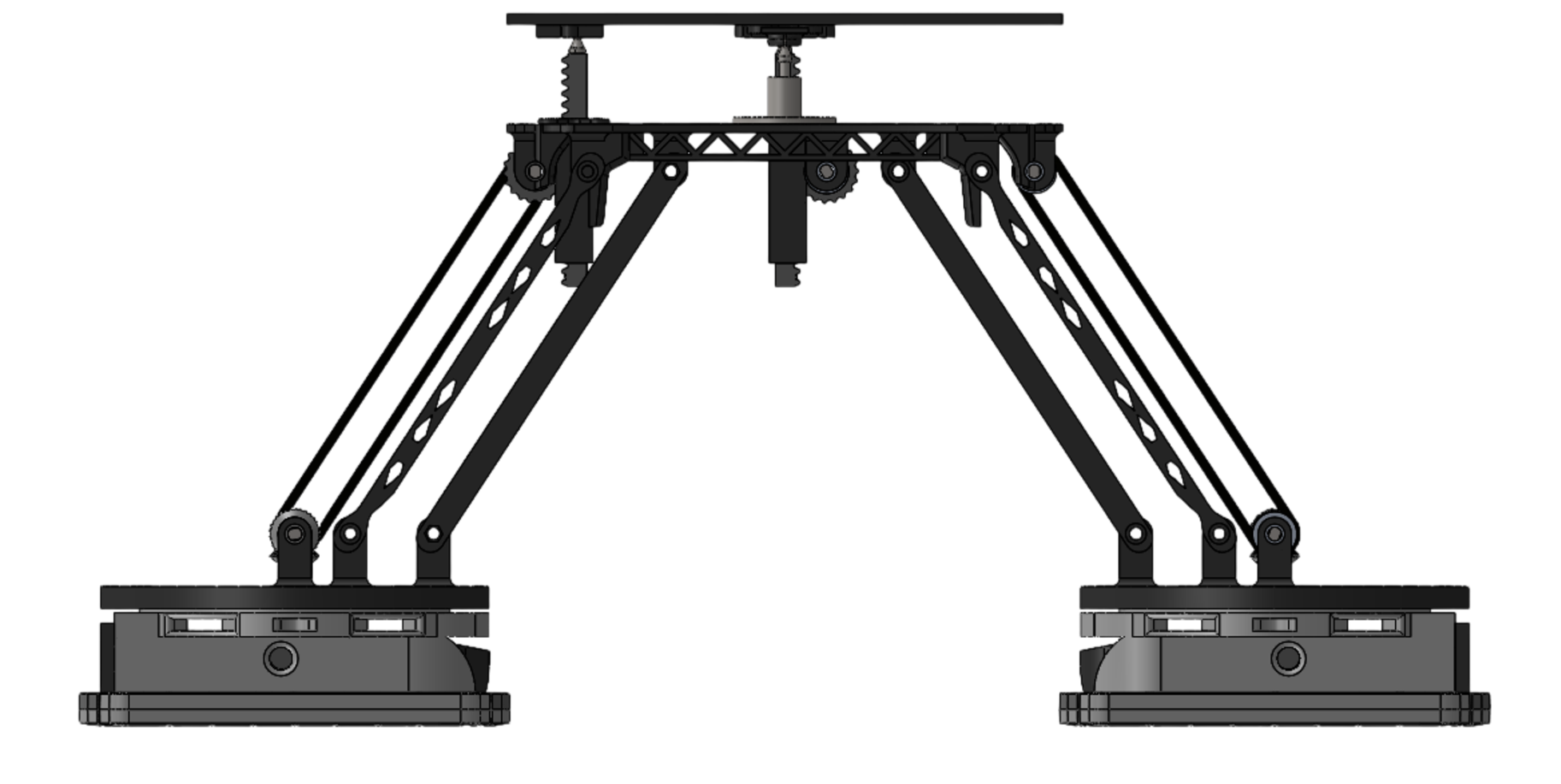}};
		
		\begin{scope}[x={(img.south east)}, y={(img.north west)}, transform shape]
			
			
			\coordinate (MoverL) at (0.189,0.12);
			\coordinate (MoverR) at (0.817,0.12);
			\coordinate (PlatMid) at (0.505,0.983);
			\coordinate (Ground)  at (0.505,0.00);
			\coordinate (MoverLbottom) at (0.19,0.075);

			\draw[colZ, very thick, <->]
			(Ground) -- node[right, inner sep=1pt, xshift=0.35mm, yshift=-1mm]{$z_p$} (PlatMid);

			\draw[colD, very thick, <->]
			($(MoverL)+(0,0.04)$) --
			node[above, anchor=south west, inner sep=1pt, xshift=4mm, yshift=0.5mm]{$d_m$}
			($(MoverR)+(0,0.04)$);
			
			\draw[colS, very thick, <->]
			(MoverLbottom) -- (0.19,0)
			node[midway, anchor=west, colS] {$z_m$};
			
			\coordinate (JbaseL) at (0.225,0.32);
			\coordinate (JtopL)  at (0.377,0.78);
			
            \draw[colI, very thick] (JbaseL) -- (JtopL) node[midway, anchor=south east, colI,  fill=white, inner sep=1pt, xshift=-0.2mm, yshift=0.2mm, font=\small] {k=156};
			
			\foreach \P in {JbaseL,JtopL}{
				\fill[white] (\P) circle (1.2pt);
				\draw[black, line width=0.4pt] (\P) circle (2pt);
			}
			
			\coordinate (MBUL) at (0.038,0.075);
			\coordinate (CUL) at (0.038, 0.32);
			\coordinate (CMU) at (0.19, 0.35);
			\coordinate (JbaseLV) at (0.225,0.35);
			\coordinate (COL) at (0.038, 0.78);
			\coordinate (TOL) at (0.038, 0.983);
			\coordinate (MO) at (0.505, 0.9);
			\coordinate (JtopLV)  at (0.377,0.9);
			
			\draw [colA, thick, <->] (MBUL) -- (CUL) node[midway, anchor=east, colA, font=\small] {$z_b=69.3$};
			\draw [colA, thick, <->] (COL) -- (TOL) node[midway, anchor=east, colA, font=\small] {$z_t=58.5$};
			\draw [colE, very thick, <->]
			(CMU) -- (JbaseLV)
			node[midway, anchor=south, colE, font=\small, fill=white, inner sep=1pt, xshift=-1mm, yshift=1.5mm] {$x_b=20$};
			\draw [colE, very thick, <->] (JtopLV) -- (MO) node[midway, anchor=south, xshift=-11.5mm, yshift=-2.9mm, colE, font=\small] {$x_t=71$};
			
			\draw[colS] (Ground) -- ++(0.5,0) -- ++(-1,0);
			\draw[colX, densely dashed] (MoverL) -- ++(0,0.2);
			\draw[colX, densely dashed] (MoverR) -- ++(0,0.2);
			\draw[colA, densely dashed] (MBUL) -- ++(0.15,0);
			\draw[colA, densely dashed] (CUL) -- ++(0.15,0);
			\draw[colA, densely dashed] (COL) -- ++(0.3,0);
			\draw[colA, densely dashed] (TOL) -- ++(0.3,0);	
			\draw[colI, densely dashed] (CMU) -- ++(0,-0.05);
			\draw[colI, densely dashed] (JbaseLV) -- ++(0,-0.05);
			\draw[colI, densely dashed] (JtopLV) -- (JtopL);
						
		\end{scope}
	\end{tikzpicture}
	
	\caption[Kinematic with marked variables]{\emph{6D-Platform MagBot} with variables \(d_m\) (mover distance), \(z_p\) (platform height), and \(z_m\) (mover flight altitude). All values are in mm.}
	\label{fig_controller_vars}
\end{figure}
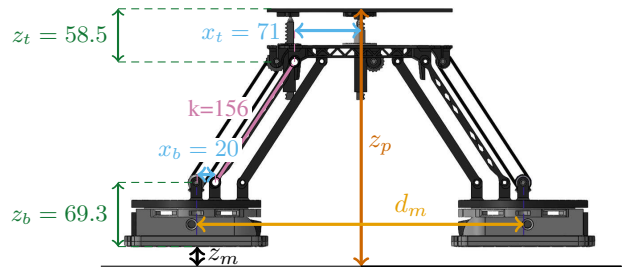
\noindent We built an inverse kinematics controller for our \emph{6D-Platform MagBot}, i.e. the controller receives the platform's set pose and outputs the x- and y-positions, as well as the $\gamma$-rotation for each of the two movers. The mover's $\alpha$- and $\beta$-rotations are always set to $0^\circ$. In the following equations, subscripts $p$ and $m$ denote the desired positions of the platform and movers, respectively. The two movers are distinguished by the subscripts $1$ and $2$, where mover $1$ is coupled to the platform's $\alpha$-rotation. The platform's z-position $z_p$ depends on the mover distance $d_m$ and the desired flight altitude of the movers $z_m$ (see Fig.~\ref{fig_controller_vars}):
\begin{equation}
    \label{eq_dm}
    d_m = 2\left( x_b + x_ t + \sqrt{k^2 - (z_p - z_m - z_b - z_t)^2}\right)
\end{equation}
where $x_b,x_t, z_b$, and $z_t$ denote known distances in the x- and z-directions (see Fig.~\ref{fig_controller_vars}). Using $d_m$, the desired mover x-,y-, and $\gamma$-positions can be calculated as follows:
\begin{align}
    x_{m,1} &= x_p - \frac{d_m}{2} \cos{(\gamma_p)}\\
    x_{m,2} &= x_p + \frac{d_m}{2} \cos{(\gamma_p)}
\end{align}
\begin{align}
    y_{m,1} &= y_p - \frac{d_m}{2} \sin{(\gamma_p)}\\
    y_{m,2} &= y_p + \frac{d_m}{2} \sin{(\gamma_p)}
\end{align}
\begin{align}
    \label{eq_cp_1}
    \gamma_{m,1} &= -s_a\cdot \alpha_p + \gamma_p\\
    \label{eq_cp_2}
    \gamma_{m,2} &= -s_b\cdot \beta_p + \gamma_p
\end{align}
where $s_a$ and $s_b$ denote gear ratios. By adding $\gamma_p$ in Eq.~\ref{eq_cp_1} and~\ref{eq_cp_2}, the $\alpha$- and $\beta$-rotations caused by the platform's $\gamma$-rotation are compensated. This inverse kinematics controller can be used with any trajectory planning approach that ensures the minimum and maximum possible distance between the two movers to protect the MagBot from damage.
\section{WORKSPACE, DYNAMICS, AND PAYLOAD}
\noindent
The \emph{6D-Platform MagBot} increases the workspace and the payload of the MagLev system compared to a single mover, as shown in Tab.~\ref{tab_workspace_payload}. Since we use a Beckhoff XPlanar system for our experiments, we compare the \emph{6D-Platform MagBot} that can be used with two APM4330-0000-0000 movers with a single mover of this type. As the tile layout of MagLev systems can be flexibly configured, the workspace in x- and y-dimensions is only limited by the number of tiles in each direction. The workspace is mainly expanded in z, $\alpha$, and $\beta$. The workspace in the $\gamma$-dimension depends on the MagLev system. As a platform $\gamma$-rotation also leads to platform $\alpha$- and $\beta$-rotations, the mover needs to compensate with its $\gamma$-rotation to keep the desired platform $\alpha$- and $\beta$-rotations. Therefore, a $360^\circ$-rotation of the platform is only possible if the MagLev system supports a $360^\circ$-rotation of the movers at each position of a tile. Using the Beckhoff XPlanar system with APS4322-0000-0000 tiles (system in our lab), the $360^\circ$-rotation of the movers is only possible at certain positions, i.e. the movers can only rotate $\pm10^\circ$ at each position of a tile. However, this is a MagLev system-specific limitation and not related to the design of the \emph{6D-Platform MagBot} that can perform a $360^\circ$ $\gamma$-rotation.\\ \\
The \emph{6D-Platform MagBot} can move a maximum payload of $2\,$kg, positioned at the center of the platform, at every platform z-position. Note that a single mover has a maximum payload of $1.8\,$kg only at a specific flight altitude of $1\,$mm. At larger flight altitudes, the payload of the mover decreases. We observed that the \emph{6D-Platform MagBot} can reliably carry the $2\,$kg payload from a mechanical perspective, but that the parameters of the low-level MagLev controller need to be carefully tuned (see Section~\ref{sec_vibration_experiments}). In addition, we can move the \emph{6D-Platform MagBot} with the full dynamics of the MagLev system, i.e. a maximum velocity of $2\,$m/s and a maximum acceleration of $10\,$m/s$^2$ using the Beckhoff XPlanar system. Therefore, based on our experience, the MagLev system itself is the limiting component in terms of payload and dynamics, but not the \emph{6D-Platform MagBot}.
\begin{table}
    \centering
    \scriptsize
    \caption{WORKSPACE AND PAYLOAD}
    \begin{tabular}{|c|c|c|c|}
        \hline
        \multicolumn{2}{|c|}{} & \textbf{Single Mover} & \textbf{6D-Platform MagBot} \\
        \hline
        \multirow{7}{*}{Workspace} & x & Limited by tile setup  & Limited by tile setup  \\
        \cline{2-4}
        & y & Limited by tile setup  & Limited by tile setup \\
        \cline{2-4}
        & z & $1\,$-$\,5\,$mm & $205\,$-$\,280\,$mm\\
        \cline{2-4}
        & $\alpha$ & $\pm5^\circ$ & $\pm14^\circ$\\
        \cline{2-4}
        & $\beta$ & $\pm5^\circ$ & $\pm14^\circ$\\
        \cline{2-4}
        & \multirow{2}{*}{$\gamma$} & $\pm10^\circ$ (everywhere) & \multirow{2}{*}{$\pm360^\circ$}\\
        \cline{3-3}
        & & $\pm360^\circ$ (at certain positions) & \\
        \hline
        \multicolumn{2}{|c|}{Max. Payload} & $1.8\,$kg & $2\,$kg\\
        \multicolumn{2}{|c|}{(Platform Center)} & (if $z=1\,$mm, otherwise less) & (at every platform z-pos)\\
        \hline
    \end{tabular}
    \label{tab_workspace_payload}
\end{table}
\section{SIMULATION}
\noindent
To enable the development of motion planning approaches for Magnetic Robotics, we developed a MuJoCo~\cite{todorov_mujoco_2012} model of the \emph{6D-Platform MagBot}, shown in Fig.~\ref{fig_sim2real}, that is integrated in the MagBotSim~\cite{bergmann_magbotsim_2025} library together with the inverse kinematics controller. MagBotSim~\cite{bergmann_magbotsim_2025} is designed specifically for Magnetic Robotics, open-source, and can be installed via PIP. In addition, MagBotSim supports common reinforcement learning (RL) APIs such as  Gymnasium~\cite{towers_gymnasium_2024} and PettingZoo~\cite{terry_pettingzoo_2021} supporting the development of intelligent motion planning approaches for MagBots, as RL is commonly used for object manipulation~\cite{iannotta_can_2025,bergmann_precision-focused_2025,stranghoner_share-rl_2025} and path planning~\cite{guan_ab-mapper_2022,damani_primal_2_2021}. Information about the \emph{6D-Platform MagBot} in simulation can be found here: \url{https://ubi-coro.github.io/MagBotSim/magbots.html}.
\section{EXPERIMENTS}
\noindent
For all experiments, we use a Beckhoff XPlanar system with APM4330-0000-0000 movers and APS4322-0000-0000 tiles, and we integrated our inverse kinematics controller in TwinCAT3. The experimental plant is configured in a $4$ tiles $\times 3$ tiles setup. The desired z-position of the movers ($z_m$ in Fig.~\ref{fig_controller_vars}) is set to $1\,$mm, as the movers are most stable at this flight altitude and can carry the maximum possible payload. For our experiments, we used the gear ratios $s_a=0.119$ and $s_b=0.131$ (Eq.~\ref{eq_cp_1},~\ref{eq_cp_2}). We estimated these values by placing an IMU chip in the center of the platform, measuring the $\alpha$- and $\beta$-rotations by rotating each mover separately in $10^\circ$-increments from $0^\circ$ to $130^\circ$, and performing linear regression using the recorded data. If not explicitly stated otherwise, we use low-level control parameter set $1$ for the mover's $\alpha$- and $\beta$-axes (see Tab.~\ref{tab_control_params}). 
\subsection{Single-Axis Positioning Accuracy}
\label{sec_sa_accuracy}
\begin{table}
    \centering
    \scriptsize
    \caption{PLATFORM POSITIONING ACCURACY}
    \begin{tabular}{|c|c|}
        \hline
        \multicolumn{2}{|c|}{\textbf{Single-Axis Movements}}\\
        \hline
        x & $0.44625\pm0.33566\,$mm\\
        \hline
        y & $0.73238\pm0.58874\,$mm\\
        \hline
        z & $0.298\pm0.29879\,$mm\\
        \hline
        $\alpha$ & $0.66504\pm0.59081\,^\circ$\\
        \hline
        $\beta$ & $0.37873\pm0.30644\,^\circ$\\
        \hline
        $\gamma$ & $0.03046\pm0.02348\,^\circ$\\
        \hline
        \multicolumn{2}{|c|}{\textbf{Multi-Axis Movements}}\\
        \hline
        x,y-Circle + Sine in z & $0.64346\pm0.27036\,$mm\\
        \hline 
        Helix & $0.73013\pm0.51699\,$mm\\
        \hline
        Cosine in $\alpha$ + Sine in $\beta$ & $1.11194\pm0.31887\,^\circ$\\
        \hline
    \end{tabular}
    \label{tab_pos_accuracy}
\end{table}
\begin{figure*}
	\centering
    \includegraphics[width=1.0\textwidth]{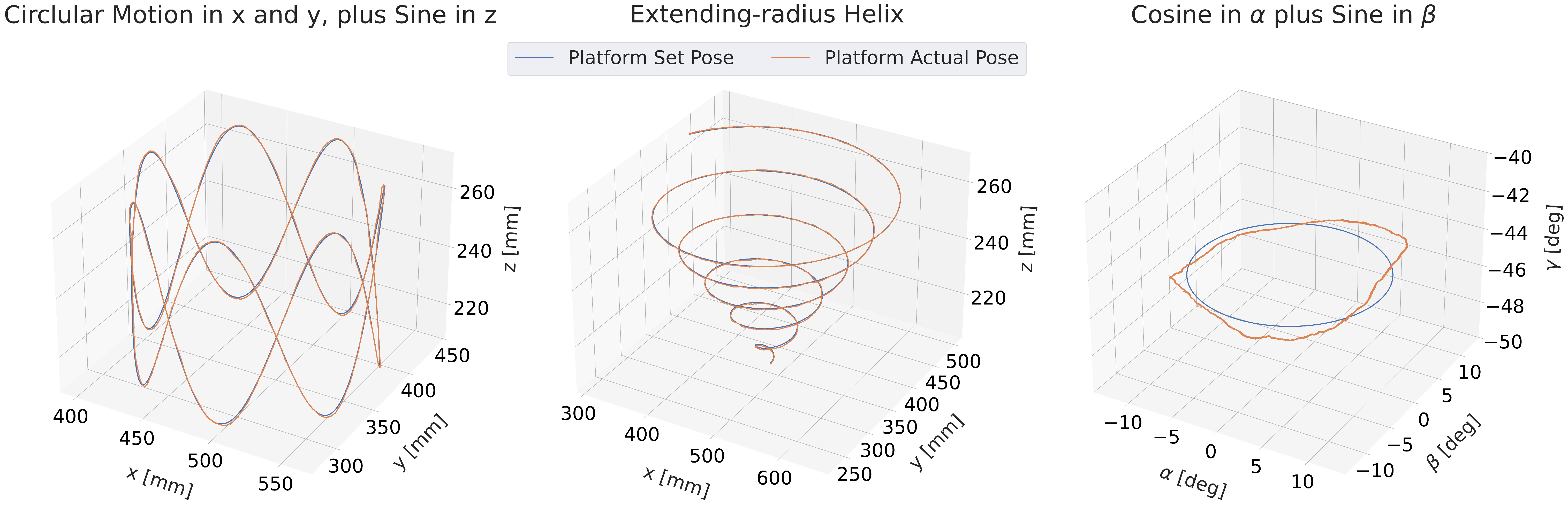}
        \caption{Multi-axis movements of the platform using our inverse kinematics controller in TwinCAT3. The platform's actual position is measured by a VICON motion capture system, while the platform's set position is directly available in TwinCAT3.}
    \label{fig_multi_axis_movements}
    \vspace{-0.5cm}
\end{figure*}
\noindent To measure the positioning accuracy of the platform using our inverse kinematics controller, we measured the actual platform pose (position and orientation) with a VICON motion capture system (MX T20 cameras, Nexus 2.6.1) and recorded the set pose of the platform that is available in TwinCAT3. In our single-axis experiments, we moved every platform axis $10$ times from its minimum to its maximum possible position, while the set positions of all other axes remain unchanged. For each platform axis, we repeated these min-max movements multiple times at different positions: x-axis with $10$ different y-positions, y-axis with $10$ different x-positions, z-axis with $25$ different x- and y-positions, $\alpha$-axis with $2$ different x-, y-, and $\gamma$-positions, $\beta$-axis with $2$ different x-, y-, and $\gamma$-positions, and $\gamma$-axis with $25$ different x- and y-positions. The maximum mover set velocity was $0.5\,$m/s and $20^\circ$/s for translational and rotational movements, respectively. Note that the measurements for the $\alpha$- and $\beta$-axes were only possible at certain positions where the $360^\circ$-rotation of the movers is available. Tab.~\ref{tab_pos_accuracy} shows the mean absolute error between the set and actual positions of the platform for each axis and the corresponding standard deviations. The accuracy in x is better than the accuracy in y, since the set y-positions of both movers are equal and thus, during motion, forces are applied perpendicular to the components depicted in orange in Fig.~\ref{fig_explosion_view}. Similarly, the accuracy in $\beta$ is better than the accuracy in $\alpha$. The reason for this result is the two toothed belts used to couple the $\alpha$-rotation of the platform with the mover's $\gamma$-rotation (see Fig.~\ref{fig_explosion_view}, red components), while only one toothed belt is used for the $\beta$-rotation. Additionally, during our experiments, we found that the platform's $\alpha$-rotation has more play than the $\beta$-rotation, resulting in a slightly delayed response. Tab.~\ref{tab_pos_accuracy} also shows that the accuracy in $\gamma$ is significantly better compared to $\alpha$ and $\beta$, since the accuracy in $\gamma$ mainly depends on the positioning accuracy of the MagLev system. These systems typically have a high positioning accuracy due to the frictionless movements~\cite{trakarnchaiyo_design_2023} and highly precise position feedback, while components such as toothed belts are involved in the case of $\alpha$ and $\beta$. In summary, our single-axis measurements reveal that the platform can be controlled with a sub-millimeter/sub-degree accuracy, which is very precise considering that all components of the \emph{6D-Platform MagBot} are low-cost and mainly 3D-printed.
\subsection{Multi-Axis Positioning Accuracy}
\label{sec_ma_accuracy}
\noindent
In addition to our single-axis experiments, we measured the positioning accuracy of the platform when moving multiple axes, again using the VICON motion capture system. We performed three different movements shown in Fig.~\ref{fig_multi_axis_movements}:
\begin{itemize}
    \item[1.] Circular motion in x and y, plus sine in z
    \item[2.] Extending-radius helix in x, y, and z (high dynamics)
    \item[3.] Cosine in $\alpha$ plus sine in $\beta$
\end{itemize}
Tab.~\ref{tab_pos_accuracy} shows that the positioning accuracy of the first movement is within the range of the single-axis accuracies in x and y, while the helix movement is close to the single-axis accuracy in y, but both motions have an accuracy in the sub-millimeter range. Note that the helix movement has significantly higher dynamics compared to the first movement (see video), explaining its slightly lower accuracy. However, since in both movements solely translational axes are moved, i.e. x, y, and z, these results are within the expected magnitude. The third movement only moves rotational axes, i.e. $\alpha$ and $\beta$. Tab.~\ref{tab_pos_accuracy} and Fig.~\ref{fig_multi_axis_movements} reveal that this movement is not as precise as the aforementioned motions, being the only one without sub-degree accuracy. Minimal inaccuracies of the 3D-printed components that lead to jerky rotational movements of the platform, the aforementioned reasons for the slightly better positioning accuracy in $\beta$ compared to $\alpha$ in the single-axis experiments, as well as position lag errors of the movers, explain this result. However, future work should investigate how to eliminate these inaccuracies, possibly by using lightweight, but more precisely manufactured components than 3D-printed ones.
\subsection{Low-level Control Parameters}
\label{sec_vibration_experiments}
\begin{figure}
	\centering
    \includegraphics[width=0.9\linewidth]{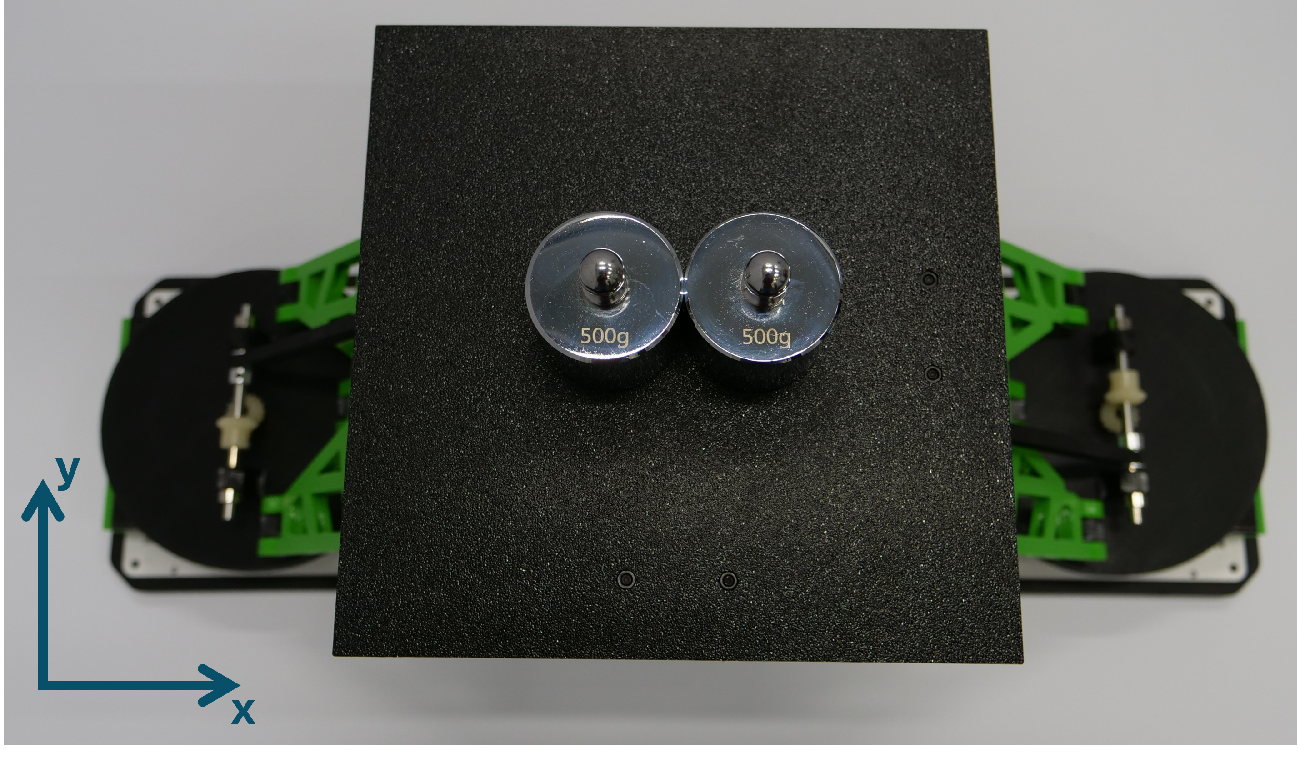}
        \caption{\emph{6D-Platform MagBot} with a payload of $1\,$kg placed in the center of the platform.}
    \label{img_magbot_with_payload}
\end{figure}

\begin{figure}
	\centering
    \includegraphics[width=\linewidth]{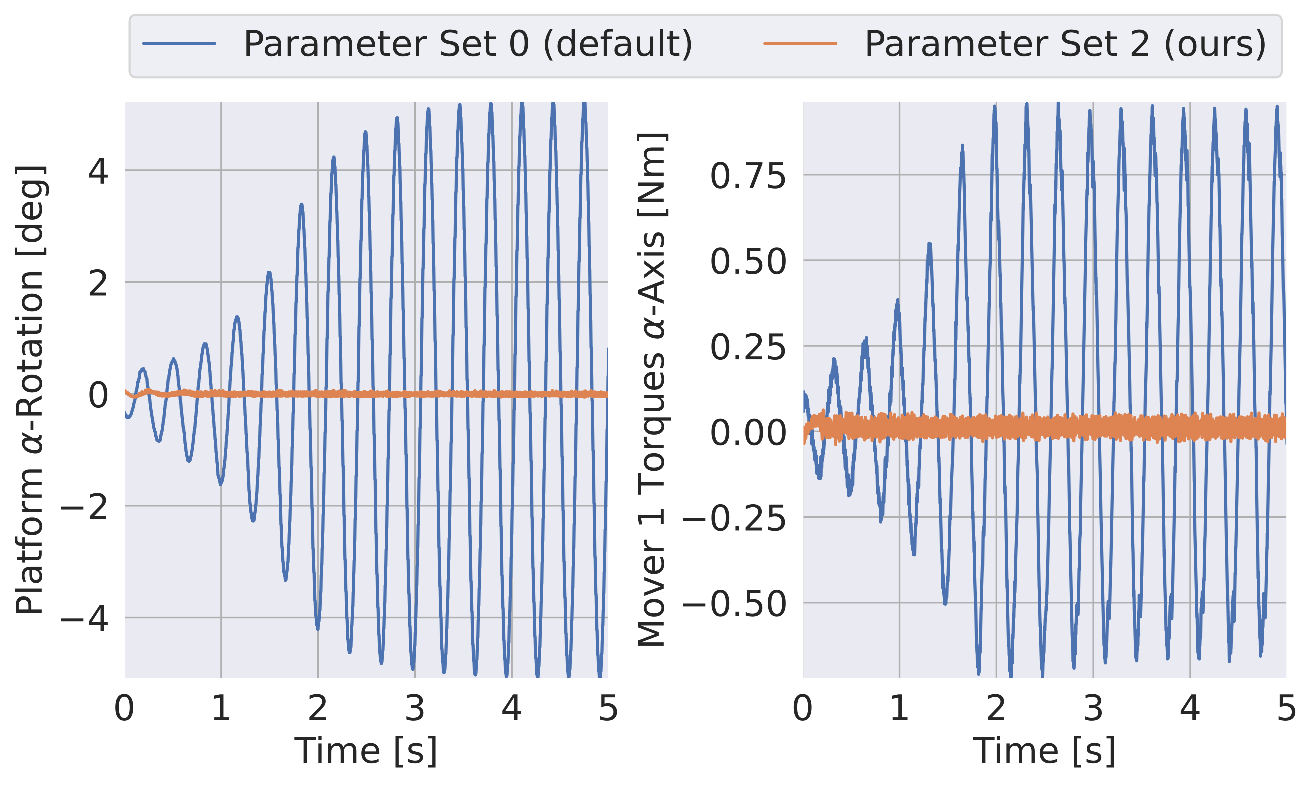}
        \caption{Platform $\alpha$-rotation (left) and mover torques (right) with $1\,$kg payload for different low-level control parameters.}
    \label{fig_vibration_1kg}
    \vspace{-0.6cm}
\end{figure}

\noindent
The XPlanar default low-level control parameters cause strong oscillations when the \emph{6D-Platform MagBot} is used both with and without a payload as soon as the movers begin to levitate. We therefore tuned the control parameters for the mover's $\alpha$- and $\beta$-axes and found two new parameter sets for the MagBot with and without payload, shown in Tab.~\ref{tab_control_params}. The parameters listed in Tab.~\ref{tab_control_params} are only used for the mover's $\alpha$- and $\beta$-axes. For all other mover axes, the XPlanar default parameters remain unchanged. To investigate the difference between the parameter sets when using a payload of $1\,$kg, we measured the platform's $\alpha$-rotation with the VICON motion capture system, as this axis is particularly sensitive to oscillations, and additionally recorded the torques that the MagLev low-level controller applies to the $\alpha$-axis of mover~{$1$}. The weights are glued to the platform, as shown in Fig.~\ref{img_magbot_with_payload}, to prevent damage to the tiles caused by falling weights. Since a payload is used, we compare the XPlanar default parameters with parameter set $2$. The results in Fig.~\ref{fig_vibration_1kg} (left) clearly show the oscillation of the platform when the default parameters are used. In contrast, when using our new parameter set, the oscillations are significantly reduced so that the system is stable. Interestingly, the same behavior is also visible in the mover's torques, shown in Fig.~\ref{fig_vibration_1kg} (right). This enables tuning the low-level control parameters without external feedback, such as a VICON system, but solely based on the data provided by the MagLev system. Thus, future work could focus on adjusting the control parameters on-the-fly, e.g. depending on the payload, based on the mover's torques.
\begin{table}[t]
    \centering
    \scriptsize
    \caption{CONTROL PARAMETER SETS FOR MOVER $\alpha$- AND $\beta$-AXES}
    \begin{tabular}{|c|c|c|c|c|c|}
        \hline
        Parameter Set & Purpose & $K_p$ & $T_n$ & $T_v$ & $T_1$ \\
        \hline 
        0 (default) & Baseline & 35.0 & 0.03 & 0.015 & 0.001 \\
        \hline
        1 (ours) & MagBot without payload & 25.0 & 0.12 & 0.04 & 0.015 \\
        \hline
        2 (ours) & MagBot with payload & 22.0 & 0.12 & 0.06 & 0.01\\
        \hline
    \end{tabular}
    \label{tab_control_params}
\end{table}
\subsection{Mover Wrenches and Payload Placement}
\begin{figure}
	\centering
    \includegraphics[width=\linewidth]{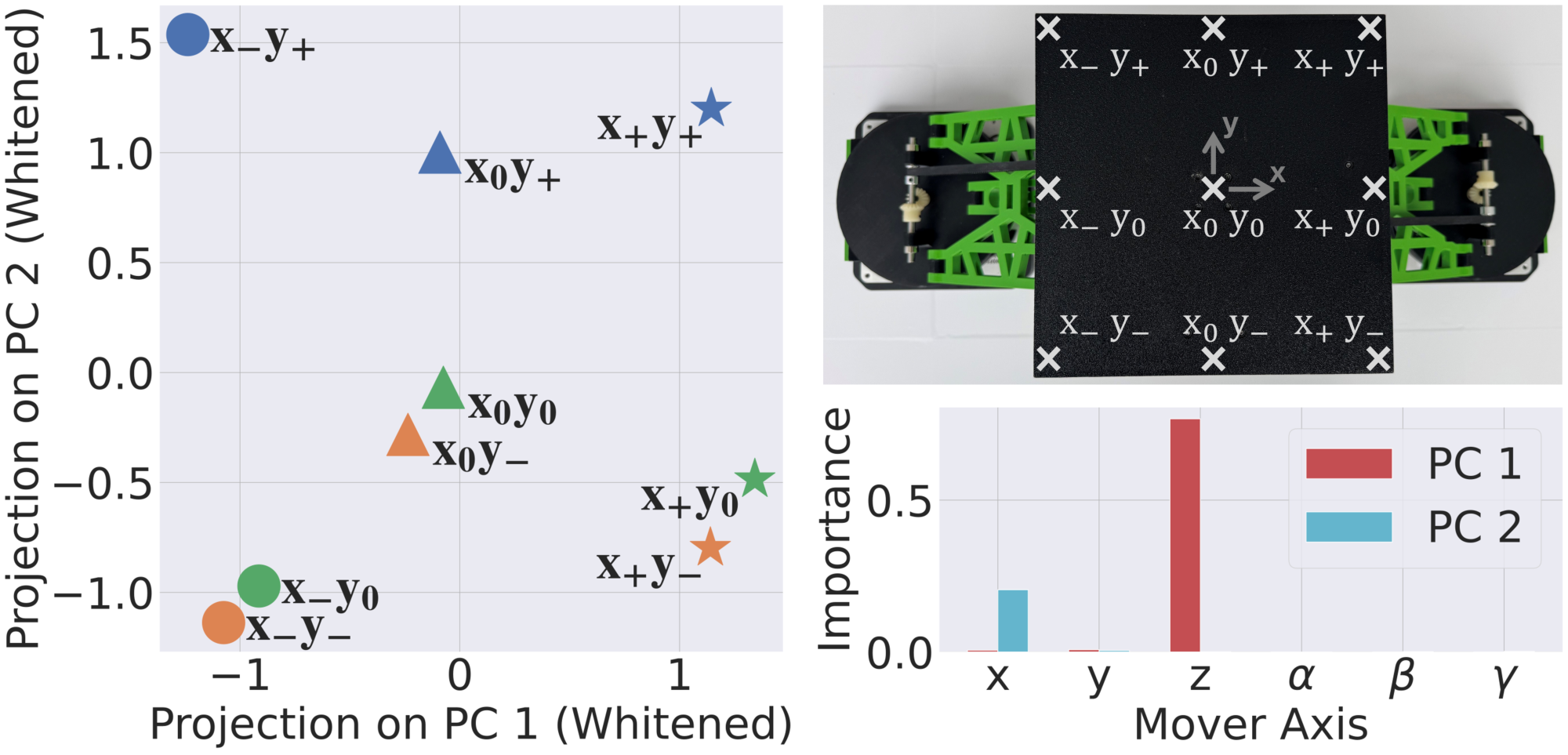}
        \caption{(Top Right) Payload positions on the platform. (Left) Projection of the differences of mean mover wrenches (see Eq.~\ref{eq_deltaF}) into a 2D-space calculated using PCA. (Bottom Right) Importance of mover wrench components for the PCA model.}
    \label{fig_payload_PCA}
\end{figure}

\noindent
Based on this result, we investigated whether we can find a pattern based on the forces and torques the MagLev low-level controller applies to the two movers that is related to the position of a payload on the MagBot's platform. To this end, we placed a $0.5\,$kg payload at the nine different positions on the platform, shown in Fig.~\ref{fig_payload_PCA} (top right), let the movers levitate at a height of $1\,$mm (no further movement) using control parameter set $2$ (see Tab.~\ref{tab_control_params}), and recorded the wrenches applied to both movers by the MagLev low-level controller. The platform's set position was set to $(480.0, 360.0, 205.0, 0.0, 0.0, 0.0)$ (in mm). For each of the nine datasets, we calculated the mean difference between the wrenches of the two movers:
\begin{equation}
    \label{eq_deltaF}
    \Delta_{\mathcal{F},i} = \frac{1}{n}\sum_{j=1}^{n}(\mathcal{F}_{j,1} - \mathcal{F}_{j,2}) \in\mathbb{R}^{6}
\end{equation}
where $i=1,...,9$ denotes the index of the dataset, $n\in\mathbb{N}$ denotes the number of samples in the dataset, and $\mathcal{F}_{j,1}\in\mathbb{R}^{6}$ and $\mathcal{F}_{j,2}\in\mathbb{R}^{6}$ denote the j-th wrench in the dataset of mover $1$ and mover $2$, respectively.
\begin{figure*}
	\centering
    \includegraphics[width=\textwidth]{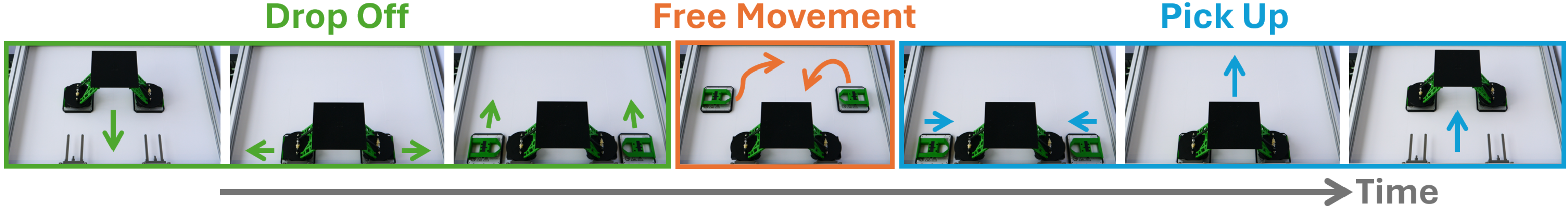}
        \caption{Visualization of the drop off and pick up process (blue) of the \emph{6D-Platform MagBot}. After the drop off, the movers can move freely (orange).}
    \label{fig_cartoon_station}
\end{figure*}
\begin{figure}
	\centering
    \includegraphics[width=0.9\linewidth]{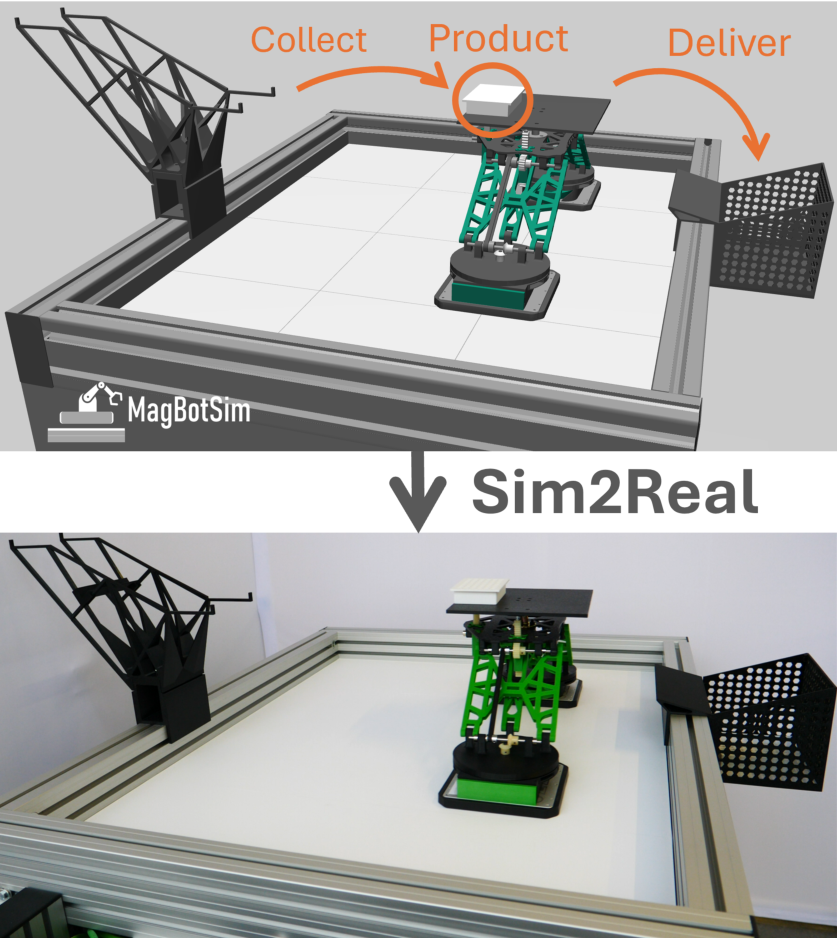}
        \caption{Application example modeled in MagBotSim~\cite{bergmann_magbotsim_2025} (top) and the real system (bottom). The task is to collect and deliver a product without additional feeding technology.}
    \label{fig_application_example}
\end{figure}
To project the data into a 2D-space, we performed a linear dimensionality reduction using principal component analysis (PCA) on $\Delta_{\mathcal{F}} = [\Delta_{\mathcal{F},1}, ..., \Delta_{\mathcal{F},9}]^{T}$, where the superscript $T$ denotes the transpose. The projected data, visualized in Fig.~\ref{fig_payload_PCA} (left), clearly shows the pattern of the payload positions on the platform. Payload positions in the second and third row are clearly distinguishable, but close to each other, as the toothed rack for the platform's $\alpha$-rotation and the ball joint on which the platform is mounted are located below the payload positions $x_0y_{-}$ and $x_0y_0$ respectively, while no toothed rack is located below $x_0y_+$. Fig.~\ref{fig_payload_PCA} (bottom right) shows that the z- and x-components of $\Delta_{\mathcal{F}}$ are most important for the PCA model, since the movers must counteract to keep the platform's z-position as soon as the payload is placed on the platform, and the platform's z-position depends on the distance between the two movers $d_m$ and the mover's flight altitude $z_m$ (see Fig.~\ref{fig_controller_vars} and Eq.~\ref{eq_dm}). The position of the payload determines which mover has to compensate more/less and thus changes the differences calculated in Eq.~\ref{eq_deltaF}. This investigation clearly shows that it is possible to find a pattern in the mover's wrenches that is related to the position of the payload, eliminating the need for an external, typically vision-based, feedback system to measure the position of the payload. Therefore, future work can use this result to estimate the position of the payload and investigate whether it is also possible to determine the payload's weight. This can be useful, e.g. for planning platform trajectories that prevent the payload from falling.
\subsection{Reconfigurability via Docking Station}
\noindent
We tested our docking station by picking up and dropping off the \emph{6D-Platform MagBot} $10$ times each, as shown in Fig.~\ref{fig_cartoon_station}. In both cases, we achieved a $100\%$ success rate, clearly demonstrating the capabilities for reconfiguration. We found that the station must be attached relatively precisely to avoid vibrations or position lag errors of the low-level mover controller. Since moving precisely into the docking station is a peg-in-a-hole task, future work should focus on applying existing algorithms, e.g.~\cite{stranghoner_share-rl_2025}, to avoid the aforementioned behavior.
\subsection{Application Example}
\noindent
Since the \emph{6D-Platform MagBot} is designed for industrial use cases, we showcase an application example where the MagBot picks up a product and delivers it into a container, as shown in Fig.~\ref{fig_application_example}. We measured the time required to pick up and deliver the product for $10$ trials. Each run was successful and took $4.298\,$s, which corresponds to a throughput of approximately $14\,$ products per minute. This application example can also be reproduced in simulation (see Fig.~\ref{fig_application_example}). Note that this task cannot be accomplished with a single mover due to the limited workspace. Additional feeding technology is commonly used in a machine to compensate for the limited workspace of MagLev systems. However, feeding technologies, such as robots, are expensive and slow compared to a MagLev mover. In contrast, \emph{6D-Platform MagBots} are low-cost, can utilize the full dynamics of the MagLev movers, and expand the task-specific capabilities of machines without requiring additional feeding technology, as emphasized in our application example. Moreover, MagBots can reduce product processing times, as product transport and processing steps can be carried out simultaneously, e.g. mixing liquids during transport. Therefore, the application of reconfigurable MagBots significantly reduces the costs and footprint of a machine, offering a cost-effective alternative to traditional machines that rely on feeding mechanisms.
\section{CONCLUSION AND FUTURE WORK}
\noindent
In this paper, we presented the \emph{6D-Platform MagBot}, a low-cost parallel kinematic with 6-DoF that couples movers of a MagLev system into a composite robotic platform. Additionally, we designed and constructed a docking station to autonomously drop off or pick up the \emph{6D-Platform MagBot}, making a machine easily reconfigurable. The \emph{6D-Platform MagBot} is lightweight, since no electronics are included and most components are 3D-printed or made of plastic. Furthermore, we integrated a MuJoCo model of the \emph{6D-Platform MagBot} in the MagBotSim library and implemented an inverse kinematics controller. Using this controller, we showed that the platform of the \emph{6D-Platform MagBot} achieves a sub-millimeter/sub-degree positioning accuracy, except for simultaneous $\alpha$- and $\beta$-rotations of the platform. Future work should investigate how to improve the latter positioning accuracy. We have also found that careful tuning of the control parameters of the low-level MagLev controller is required to avoid oscillations. Furthermore, we showed that no external feedback, e.g. a VICON system, is required to tune these parameters, as the oscillations are also clearly visible in the wrenches applied to the mover by the MagLev controller. Moreover, we demonstrated that it is possible to find a pattern in the wrenches of the MagLev controller that is related to the position of a payload on the MagBot's platform. \emph{6D-Platform MagBots} provide multiple future research directions. Firstly, since \emph{6D-Platform MagBots} couple movers into composite robotic platforms, it breaks the classic separation between transportation and manipulation in standard multi-agent systems, leaving a gap for new motion planning approaches that also consider the high-dynamics of MagLev systems. Secondly, these motion planning approaches must additionally provide strong safety guarantees to enable the use of MagBots in industrial applications. Thirdly, human-robot interaction and collaboration, including topics such as reliable user recognition, easy instructability, smooth and safe handover interactions, and the use of imitation learning to acquire human-aligned behaviors, are largely unexplored for MagLev systems and MagBots. In conclusion, the \emph{6D-Platform MagBot} significantly expands the reachable workspace, payload, and functional dexterity of a MagLev system, enabling the unification of transportation and manipulation, and therefore making manufacturing systems simultaneously more agile, efficient, and flexible.



\bibliographystyle{IEEEtran}
\bibliography{References6DPlatformMagBot}

\end{document}